\documentclass{sig-alternate-05-2015}
\pdfoutput=1 

\usepackage[    
protrusion=true, 
expansion=true,
tracking=true,
kerning=true,
spacing=true,
letterspace=50,
shrink=40,      
factor=1000
]    
{microtype}

\usepackage{graphicx}
\usepackage{amssymb}
\usepackage{amsmath}
\usepackage{multirow}
\usepackage{rotate}
\usepackage{subfigure}
\usepackage{epstopdf}
\usepackage{tabularx}
\usepackage{booktabs}
\usepackage{paralist}
\usepackage{graphicx}
\usepackage{ifpdf}
\usepackage{mathdots}
\usepackage{lscape}
\usepackage{setspace}
\usepackage{relsize}
\usepackage{amsfonts}
\usepackage{mathrsfs}
\usepackage{sidecap}
\usepackage[table]{xcolor}

\definecolor{verylightgreen}{RGB}	{204,255,204}
\definecolor{verylightred}{RGB}		{255,204,204}

\definecolor{verylightyellow}{RGB}		{255,255,204}
\renewcommand{\tabcolsep}{4pt}

\usepackage{amssymb}
\usepackage{amsmath}
\usepackage{mathdots}
\usepackage{mathtools}
\usepackage{lscape}
\usepackage{listings}
\usepackage{algorithm}
\usepackage{algorithmicx}
\usepackage{algpseudocode}
\algnotext{EndFor}
\algnotext{EndIf}
\algnotext{EndWhile}
\algnotext{EndProcedure}

\usepackage{wrapfig}
\usepackage{xspace}
\usepackage{xcolor}

\newcommand{\etal}{\emph{et al.}\xspace}

\definecolor{gray}{RGB}{20,20,20}
\definecolor{greencm}{RGB}{0,153,0}

\usepackage{pifont}

\newcolumntype{H}{>{\setbox0=\hbox\bgroup}c<{\egroup}@{}}

\definecolor{verylightgreennew}{RGB}	{220,255,220}
\definecolor{verylightrednew}{RGB}		{255, 230, 230}
\definecolor{verylightrednewlighter}{RGB}		{255, 229, 239}
\definecolor{lightgraynew}{rgb}{0.9,0.9,0.9}

\definecolor{gray}{RGB}{150,150,150}
\definecolor{theblue}{RGB}{0,0,180}
\usepackage{mathtools}

\definecolor{theblue}{RGB}{0,0,180}
\colorlet{TufteRed}{red!80!black}
\colorlet{thered}{TufteRed}

\newcommand{\be}{\begin{equation}}
\newcommand{\ee}{\end{equation}}
\newcommand{\bea}{\begin{eqnarray}}
\newcommand{\eea}{\end{eqnarray}}
\newcommand{\bit}{\begin{itemize}}
\newcommand{\eit}{\end{itemize}}

\definecolor{lightgray}{rgb}{0.93,0.93,0.93}

\definecolor{lightblue}{rgb}{0.5,0.90,1.0}
\definecolor{lightgreen}{rgb}{0.5,0.92,0.5}
\definecolor{lightred}{rgb}{0.98,0.5,0.5}
\definecolor{lightyellow}{rgb}{1,0.90,0.40}

\definecolor{lightgray}{rgb}{0.93,0.93,0.93}
\definecolor{lightblue}{rgb}{0.5,0.90,1.0}
\definecolor{lightgreen}{rgb}{0.5,0.92,0.5}
\definecolor{lightred}{rgb}{0.98,0.5,0.5}
\definecolor{lightyellow}{rgb}{1,0.90,0.40}

\definecolor{mygreen}{rgb}{0,0.6,0}
\definecolor{mygray}{rgb}{0.5,0.5,0.5}
\definecolor{mymauve}{rgb}{0.58,0,0.82}
\definecolor{subtleblue}     {rgb}{0.02,0.04,0.48}
\definecolor{subtlered}      {rgb}{0.65,0.04,0.07}
\definecolor{subtlegreen}    {rgb}{0.06,0.44,0.08}
\definecolor{subtledarkblue} {rgb}{0,.1,.6}
\definecolor{lightsubtleblue}{rgb}{0,.4,.6}
\definecolor{ecru}           {rgb}{1.0,.98823,.95686}

\newcommand\TTT{\rule{0pt}{3.2ex}}
\newcommand\BBB{\rule[-1.4ex]{0pt}{0pt}}

\newcommand\TT{\rule{0pt}{2.3ex}}
\newcommand\BB{\rule[-1.0ex]{0pt}{0pt}}
\newcommand\T{\rule{0pt}{3.2ex}}

\newtheorem{Definition}{Definition}[section]

\usepackage{booktabs}
\algblockdefx[parallel]{parfor}{parend}
[1][]{\textbf{parallel for} #1}
{\textbf{end parallel}}

\algblockdefx[fornew1]{foreach}{endforeach}
[1][]{\textbf{for} #1}
{\textbf{end for}}

\newcommand{\eat}[1]{}
\algrenewcommand{\alglinenumber}[1]{\fontsize{6.5}{7}\selectfont#1}
\providecommand{\multiline}[1]{\State \parbox[t]{\dimexpr\linewidth-\algorithmicindent}{#1\strut}}

\usepackage{comment}
\usepackage{amssymb}
\usepackage{amsmath}
\usepackage{mathdots}
\usepackage{lscape}
\makeatletter
\def\vcdots{\vbox{\baselineskip4\p@ \lineskiplimit\z@
    \kern3\p@\hbox{.}\hbox{.}\hbox{.}\kern3\p@}}
\makeatother

\newcommand{\bigO}[1]{\ensuremath{\mathcal{O}(#1)}}

\newcommand{\rsm}{\ensuremath{\textsc{rsm}}}
\newcommand{\wvrn}{\ensuremath{\textsc{wvrn}}}
\newcommand{\ml}{\ensuremath{\textsc{ml}}}
\newcommand{\rml}{\ensuremath{\textsc{rml}}}
\newcommand{\irml}{\ensuremath{{\mathit{i}\textsc{rml}}}}
\newcommand{\data}[2]{{\mathsf{#1}\text{--}{\mathsf{#2}}}} 
\usepackage{epsfig}
\usepackage{amssymb}
\usepackage{amsmath}
\usepackage{amsfonts}

\usepackage{comment}
\usepackage{eqparbox}

\renewcommand{\subsubsection}[1]{\medskip\noindent\textbf{{{#1}:\,}}}

\colorlet{TufteRed}{red!80!black}
\definecolor{theblue}{RGB}{0,0,180}
\colorlet{thered}{TufteRed}

\usepackage{hyperref}
\hypersetup{colorlinks=true,urlcolor=theblue,citecolor=theblue,linkcolor=theblue}

\usepackage{amsfonts}
\usepackage{ragged2e}
\usepackage{setspace}
\graphicspath{{./}{./images/}}
\usepackage{upgreek}

\usepackage[notheorems]{rr-math}

\usepackage{nicefrac}
\usepackage{mathtools}

\newcommand{\eg}{\emph{e.g.}}
\newcommand{\ie}{\emph{i.e.}}
\newcommand{\iid}{\emph{iid}}

\providecommand{\multiline}[1]{\State \parbox[t]{\dimexpr\linewidth-\algorithmicindent}{#1\strut}}

\renewcommand{\Pr}{\mathbb{P}} 
 
\renewcommand{\Std}{\mathbb{S}}

\newcommand{\Touter}{\tau_{\max}}

\RequirePackage{mathtools}

\newcommand{\ds}\displaystyle
\newcommand{\mbb}\mathbb
\newcommand{\mc}\mathcal
\newcommand{\del}\nabla

\newcommand{\beqstar}{\begin{eqnarray*}}
\newcommand{\eeqstar}{\end{eqnarray*}}

\definecolor{thegreen}{rgb}{0,.5,0}
\definecolor{idea}{rgb}{0,.6,0.1}
\definecolor{problem}{rgb}{0.7,0,0.1}
\definecolor{comment-green}{rgb}{0,.3,0}
\definecolor{theblue}{rgb}{0,0,.8}
\definecolor{light-gray}{gray}{0.98}
\definecolor{comment-color}{rgb}{0,0,.8}
\definecolor{string-color}{rgb}{0,.75,0}
\definecolor{border-blue}{rgb}{0,0,.6}

\usepackage{setspace}

\graphicspath{{./}{./images/}}
\newcommand{\all}{\ensuremath{{:}}}
\renewcommand{\th}{\ensuremath{\rm th}}

\renewcommand{\T}{\ensuremath{\top}}

\providecommand{\mc}{\ensuremath{\omega}}

\providecommand{\p}{\ensuremath{p}}

\renewcommand{\E}{\ensuremath{{E}}}

\newcommand{\A}{\ensuremath{\mA}}

\newcommand{\simf}{\mathsf{\bf S}}

\renewcommand{\P}{\ensuremath{\mP}}

\newcommand{\n}{\ensuremath{n}} 
\newcommand{\m}{\ensuremath{m}} 
 
\newcommand{\f}{\ensuremath{d}}
\renewcommand{\d}{\ensuremath{d}} 
\renewcommand{\ds}{\ensuremath{\d^{\prime}}} 
\renewcommand{\k}{\ensuremath{k}} 

\providecommand{\Vl}{\ensuremath{V^{\ell}}} 
\providecommand{\Vu}{\ensuremath{V^{u}}}

\renewcommand{\O}{\ensuremath{\mathcal{O}}}

\newcommand{\N}{\ensuremath{\Gamma}}

\usepackage{booktabs}
\algblockdefx[parallel]{parfor}{parend}
[1][]{\textbf{parallel for} #1}
{\textbf{end parallel}}

\algrenewcommand{\alglinenumber}[1]{\scriptsize#1}
\algrenewcommand{\alglinenumber}[1]{\fontsize{6.5}{7}\selectfont#1}
\algrenewcommand{\alglinenumber}[1]{\fontsize{6.5}{7}\selectfont#1}
\providecommand{\multiline}[1]{\State \parbox[t]{\dimexpr\linewidth-\algorithmicindent}{#1\strut}}

\usepackage{comment}
\usepackage{amssymb}
\usepackage{amsmath}
\usepackage{mathdots}
\usepackage{lscape}
\makeatletter
\def\vcdots{\vbox{\baselineskip4\p@ \lineskiplimit\z@
    \kern3\p@\hbox{.}\hbox{.}\hbox{.}\kern3\p@}}
\makeatother

\renewcommand{\bigO}[1]{\ensuremath{\mathcal{O}(#1)}}
\usepackage{ragged2e}
\usepackage{epsfig}
\usepackage{amssymb}
\usepackage{amsmath}
\usepackage{amsfonts}

\usepackage{comment}
\begin{document}
\setcopyright{rightsretained}
\conferenceinfo{\hspace*{-0.5mm}MLG KDD}{2016 San Francisco, CA USA}%
\CopyrightYear{2016}
\title{Relational Similarity Machines}

\numberofauthors{3} 
\author{
\alignauthor
Ryan A. Rossi\\
       \affaddr{Palo Alto Research Center}\\
       \email{rrossi@parc.com}
\alignauthor
Rong Zhou\\
       \affaddr{Palo Alto Research Center}\\
       \email{rzhou@parc.com}
\alignauthor
Nesreen K. Ahmed\\
       \affaddr{Intel Labs}\\
       \email{nesreen.k.ahmed@intel.com}
}

\maketitle
\begin{abstract}
This paper proposes Relational Similarity Machines (RSM): a fast, accurate, and flexible relational learning framework for supervised and semi-supervised learning tasks. Despite the importance of relational learning, most existing methods are hard to adapt to different settings, due to issues with efficiency, scalability, accuracy, and flexibility for handling a wide variety of classification problems, data, constraints, and tasks.  For instance, many existing methods perform poorly for multi-class classification problems, graphs that are sparsely labeled or network data with low relational autocorrelation. In contrast, the proposed relational learning framework is designed to be (i) fast for learning and inference at real-time interactive rates, and (ii) flexible for a variety of learning settings (multi-class problems), constraints (few labeled instances), and application domains. The experiments demonstrate the effectiveness of RSM for a variety of tasks and data.
\end{abstract}


\section{Introduction} \label{sec:intro}
\noindent
Networks (relational data, graphs) encode dependencies between entities (people, computers, proteins) and allow us to study phenomena across social~\cite{ahmed2012space}, technological~\cite{mahadevan2006internet}, and biological domains~\cite{bassett2006small}.
Recently, \underline{r}elational \underline{m}achine \underline{l}earning ($\rml$) methods were developed to leverage relational dependencies~\cite{introSRL07,de2008probILP,rossi2012transforming} between nodes to improve predictive performance~\cite{wvrn,Friedman99:PRMs,mcdowell2009cautious,mcdowell2010meta,de2008probILP,Neville03simpleestimators}.

Relational classifiers can sometimes outperform traditional $\iid$ $\ml$ techniques by exploiting dependencies between class labels (attributes) of related nodes.
However, the performance of $\rml$ methods can degrade when there are few labeled instances (majority of neighboring instances are also unlabeled).
Collective Classification (CC) aims to solve this problem by iteratively predicting labels and propagating them to related instances~\cite{sen2008collective}.
Unfortunately, the performance of CC methods may also degrade when there are very few labels available (e.g., label density $<$ $0.01$)~\cite{mcdowell2009cautious}.
In both situations, if not careful, the performance of $\rml$ methods may degrade to a point where $\iid$ techniques perform better.

\begin{figure*}[t!]
\centering
\includegraphics[width=0.76\linewidth]{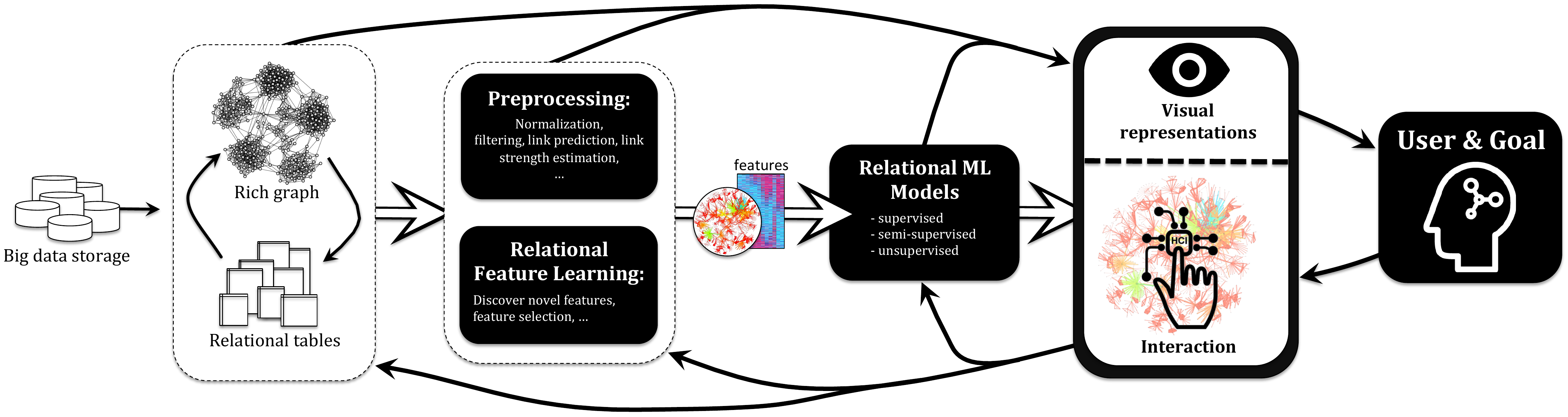}
\vspace{2mm}
\caption{Interactive Relational Machine Learning (iRML) Paradigm. Visual encoding of the results from the various components are indicated via the top arrows, whereas user interactions are represented by the arrows at the bottom.}
\label{fig:irml-framework}
\end{figure*}

Despite the fundamental importance of these techniques, the vast majority of $\rml$ methods rely on a significant amount of relational autocorrelation (homophily~\cite{mcpherson2001homophily}) existing in the data.
It has been noted that $\rml$ techniques may perform worse than $\iid$ methods when there is low or even modest relational autocorrelation.
In addition, existing methods also have difficulty learning with graph data that is large, noisy, probabilistic, sparsely labeled, attributed, 
and are sensitive to many other issues and data characteristics that often arise in practice.
Moreover, obtaining labeled data is also expensive, and thus $\rml$ methods should be robust to learning with few labeled instances.
Furthermore, relational representation and transformations of the nodes, links, and/or features can dramatically affect the capabilities and results of such algorithms~\cite{rossi2012transforming}.
In general, $\rml$ methods are sensitive to many other important and fundamental issues that arise in practice (and thus unique to interconnected and dependent relational data).

To solve these problems, this paper introduces relational similarity machines ($\rsm$) --- an efficient, flexible, and highly accurate relational learning framework.
$\rsm$ is based on the fundamental notion of similarity and is well-suited for classification and regression tasks in arbitrary networked data.
It also generalizes to both graph-based supervised and semi-supervised learning (SSL)~\cite{zhu2009introduction} and gives rise to a variety of methods for both settings.
In particular, $\rsm$ is extremely fast, accurate, and flexible with many interchangeable components.
$\rsm$ learning and inference is fast, space-efficient, and highly scalable for \emph{real-time interactive learning}.
In addition, $\rsm$ is designed to be intuitive and easy to adapt and encode application constraints.
Moreover, $\rsm$ has many other attractive properties including its robustness to noisy links, nodes, and attributes, as well as graph data with only a few labeled instances and attributes.

Existing methods perform poorly for data that exhibits low relational autocorrelation, which occurs frequently in practice.
In contrast, $\rsm$ naturally handles relational data with varying levels of autocorrelation by adjusting a simple hyperparameter.
In particular, the key contributions of this work as well as the advantages of $\rsm$ over existing methods are as follows:
{
\smallskip
\begin{compactenum}[{$\vcenter{\hbox{\tiny$\bullet$}}$\leftmargin=0em\itemsep=0pt}]
\itemsep 3pt
\item A general principled relational learning framework is proposed for large-scale supervised and semi-supervised learning (SSL) in multi-dimensional networks that are noisy, sparsely labeled, and contain only modest or even low levels of relational autocorrelation.
\item Naturally generalizes for large multi-class problems
\item Flexible with many interchangeable components (\eg, similarity function, collective inference)
\item Fast learning and inference methods (for applications requiring real-time performance). 
\item Easily parallelizable for shared- and distributed-memory architectures with speedups that are nearly linear.
\item The contribution of the relational information can be learned from the data directly (or quickly tuned by the user).
\item Effective for sparsely labeled graph data as well as situations where the labels of the training data are skewed towards one or more class (\eg, represented disproportionately).
\item $\rsm$ generalizes across the space of potential classifiers including simple approaches that leverage non-relational $\iid$ data, to those that leverage the graph topology (and possibly a set of attributes), as well as classifiers that leverage both. In fact, these are all special cases that arise under certain conditions (hyper-parameters combinations). 
Further, a collective inference approach is derived for the space of classifiers that leverage graph topology in some fashion (where $G$ may be given as input or perhaps created, $\eg$ using SSL).
\item $\rsm$ is robust to networks with modest or even low levels of relational autocorrelation
\item Excellent predictive accuracy and performs significantly better than existing approaches with comparable time and space constraints.
\item Amenable for streaming and online settings with efficient updates, learning, and inference.
\end{compactenum}
\medskip
\noindent
Note that $\rsm$ is also useful for heterogeneous networks and leverages both link and node attributes.
}

\section{Interactive RML}\label{sec:irml}
\noindent
Relational Machine Learning ($\rml$)~\cite{introSRL07} methods exploit the relational dependencies between nodes to improve predictive performance~\cite{wvrn}.
However, these approaches often fail in practice due to low relational autocorrelation, noisy links, sparsely labeled graphs, and data representation~\cite{rossi2012transforming}.

\subsection{iRML Paradigm}
\noindent
To overcome these problems, the Interactive Relational Machine Learning ($\irml$) paradigm was envisioned~\cite{rossi2016aaai} in which users interactively specify relational models and data representation (via transformation techniques for the graph structure and features), as well as perform evaluation, analyze errors, and make adjustments and refinements in a closed-loop~\cite{rossi2016aaai}.
In this work, humans interact with relational learning algorithms by providing input (in the form of labels, similarity/kernel function, hyper-parameters, priors, confidence/uncertainty about particular instances, learning rate, corrections, rankings, probabilities, evaluation) while observing the output (in the form of predictions, uncertainty, feedback, and any useful visual representation of the data).

{
\smallskip
\noindent
Desiderata for the $\irml$ paradigm are as follows:
\smallskip
\begin{compactitem}[{$\vcenter{\hbox{\tiny$\bullet$}}$\leftmargin=0em\itemsep=0pt}]
\itemsep 1pt
\item \textbf{Immediate visual feedback}.
$\irml$ methods should be optimized for the way humans learn~\cite{ahlberg1992dynamic,thomas2005illuminating,ahmed-icwsm15}.
Thus, they should provide \emph{immediate} and \emph{continuous visual feedback} upon every interaction (e.g., change of a slider for filtering uncertain or misclassified nodes, selection of a subgraph for modeling, or correcting the class label of a node).
Further, interactive queries need to be rapid, incremental and reversible with immediate visual feedback.

\item \textbf{Flexibility and generality}.
Methods should be useful for a wide variety of data, constraints, and learning scenarios. 
They need to also be robust for learning sparsely labeled graphs, noisy relational data, low and varying levels of relational autocorrelation, and other problems that frequently arise in practice.

\item \textbf{Effectiveness}.
Models must have good predictive quality with low error, variance, and bias. 

\item \textbf{Scalable methods}. 
Fast time- and space-efficient learning methods capable of interactive rates is an important and key requirement.
The requirement of rapid and interactive model updates often dictates trading off accuracy with speed.
Thus, network sampling methods may be used to balance speed and accuracy.

\item \textbf{Accessibility and simplicity}. 
To be accessible to domain but non-ML experts, $\irml$ methods must be carefully designed to be simple, intuitive, and easy-to-use.
Whenever possible, assistance and guidance from the system is desired.

\item \textbf{Principled models}. 
Another challenge is the design of intuitive learning and inference methods to facilitate interactive reasoning, understanding, and derivation of theoretical behavior and guarantees. 
This enables quick understanding and refinement by the user, while also providing a means to backtrack if warranted to understand a specific outcome or anomaly.

\item \textbf{Unified \& expressive models}. 
A unifying family of relational learning methods that express a large and multi-faceted space of relational models. 
These models must perform well across a variety of different data, characteristics, and assumptions.
They must also generalize to a variety of learning settings (e.g., relational active learning, online/incremental learning).
\end{compactitem}
}

\subsection{Visual Representation and Interaction}
\noindent
Visual analytic techniques are also developed that combine a wide variety of visual representations and interactive techniques to exploit human capabilities for seeing, exploring, and understanding large amounts of information. 
A few of the features/advantages are given below:
\smallskip
\begin{compactitem}[{$\vcenter{\hbox{\tiny$\bullet$}}$\leftmargin=0em\itemsep=0pt}]
\itemsep 0pt
\item After each graph manipulation or user interaction (\eg, via visual interactions, change of a slider), the $\irml$ method is updated immediately with visual feedback after each change.
For instance, suppose a user finds that a node is labeled incorrectly via the visual analytic techniques, and simply corrects the error immediately by simply clicking the node and updating its field.
Immediately after the correction is made, all features, models, and visual representations that depend on that value are immediately updated on the fly.
Hence, after each graph manipulation (or user interaction) such as inserting, filtering, or permanently deleting nodes and links.

\item Select or hover over nodes and edges to analyze their class label, uncertainty, estimated class label probabilities, as well as other properties, attributes, relational features, as well as topology features such as graphlet statistics~\cite{ahmed2016kais,ahmed2015icdm}.

\item Real-time interactive visual graph querying \emph{and} filtering capabilities, \eg, filter all uncertain nodes above a user-specified threshold, or select all misclassified nodes for further analysis. 

\item Color \emph{and} size of nodes and edges can be visually encoded as a function of measures derived from learning and inference such as node (edge) labels, uncertainty, or a learned feature or meta-feature, among other possibilities. Non-relational attributes and network properties derived from the topology may also be used.

\item Multiple potentially disconnected subgraphs may be selected for learning, \eg, by brushing over interesting regions of the network visually or by sliders that control filtering/selection, etc. 
\eat{Multiple selections from different regions of the graph are also supported.}
\item Nodes (or edges) can be labeled easily, \eg, double-clicking a node of interest (or set of nodes after selection). 

\end{compactitem}
\smallskip
\noindent
Nearly all visualizations are interactive and support brushing, linking, zooming, panning, tooltips, etc.
Multiple visual representations of the graph data are supported, including the multi-level graph properties (e.g., interactive scatter plot matrix, and other statistical plots).
Dynamic network analysis and visualization tools to understand the temporal dependencies to better leverage them in learning.
Network may also be searched via textual query (e.g., node name).
There are many other features including full customization of the visualization (color, size, opacity, background, fonts, etc), text annotation, graph layouts, collision detection, fish eye, and many others.
See~\cite{rossi2016aaai} for more details.
Note $\rsm$ was implemented in the interactive graph mining and visualization platform proposed in~\cite{ahmed-icwsm15}.

\section{Relational Similarity Machines} \label{sec:method}
\noindent
Given an attributed graph $G=(V,E,\mX,Y,C)$ where 
$V$ is a set of $\n=|V|$ nodes (\eg, IoT devices in a sensor network), 
$E$ is a set of $\m=|E|$ edges (\eg, communications between devices).
We define $\mX \in \RR^{\n \times \f}$ as a matrix where rows represent nodes and columns are features. 
Further, let $\vx_{i}$ be the feature vector for $v_i \in V$.
For clarity, we also use $\mX_{i \all}$ for the $i^{\th}$ row of $\mX$ and $\mX_{\all k}$ to denote the $k^{\th}$ column of $\mX$.
\begin{Definition}
Given a graph $G$, a known set of node labels $Y^{\ell} = \{y_i | v_i \in V^{\ell}\}$ for nodes $V^{\ell} \subset V$, 
the within-network classification task is to infer $Y^{u}$ --- the set $V^{u} = V \setminus V^{\ell}$ of remaining vertices with unknown labels.
\end{Definition}
The set of class labels is denoted as $C$ and $\k=|C|$ is the number of unique labels.
Let $\A = \big[ a_{ij}\big]$ be the adjacency matrix of $G$ where $A_{ij}=1$ if there exists $(v_i,v_j) \in E$ and $A_{ij}=0$ otherwise.
Furthermore, $\A$ may encode an edge attribute, that is, $A_{ij}=\Phi_{ij}$.

{
\algblockdefx[parallel]{ParFor}{EndPar}[1][]{$\textbf{parallel for}$ #1 $\textbf{do}$}{}
\algnotext{EndPar}
\begin{figure}[t!]
\begin{center}
\begin{minipage}{1.0\linewidth}
\begin{algorithm}[H]
\centering
\caption{\,\fontsize{8}{9}\selectfont Relational Similarity Machines (RSM). A general and flexible relational learning framework based on the notion of maximum similarity.}
\label{alg:rsm}{
\newcommand{\setAlgFontSize}{\fontsize{7.5pt}{8pt}\selectfont} 
\renewcommand{\multiline}[1]{\State \parbox[t]{\dimexpr\linewidth-\algorithmicindent}{#1\strut}}
\newcommand{\multilinenospace}[1]{\State \parbox[t]{\dimexpr0.9\linewidth-\algorithmicindent}{\begin{spacing}{1.1}\setAlgFontSize#1\strut \end{spacing}}}
\begin{spacing}{1.4}
\algrenewcommand{\alglinenumber}[1]{\fontsize{6}{7}\selectfont#1:\,}
\setAlgFontSize
\begin{algorithmic}[1]
\vspace{1mm}
\State Normalize all data (if needed)
\label{alg:normalize-train-nodes-feature-vector}

\State Generate an ordering of $V^{u}$ \label{alg:order-nodes}

\multiline{Estimate the class priors $\mP = \big[\begin{smallmatrix} \cdots & \vp_i & \cdots \end{smallmatrix}\big]^{\T}$ where $\vp_i \in \RR^{\k}$ is the estimated prior for $v_i$ (see Section~\ref{sec:class-prior-estimation}).
\label{alg:rsm-estimated-prior}}

\multiline{Compute graph topology features using $G=(V,E)$ (\eg, based on k-graphlets where $k=\{3,4,...\}$~\cite{ahmed2015icdm}, among other important and efficient graph properties) and append these to $\mX$ and $\mZ$
\label{alg:initial-graph-topology-features}}

\State Set $\tau \leftarrow 1$
\label{alg:curr-iteration-time}

\Repeat \label{alg:rsm-conv-collective-inference} \Comment{outer iterations $\tau = 1,2,...,\Touter$}

\multiline{Compute relational features based on \emph{neighbor classes}\label{alg:rsm-rel-feat-neigh-labels}}
\multiline{Compute relational features based on \emph{neighbor attributes}\label{alg:rsm-rel-feat-neigh-attrs}}
\multiline{Append the relational features from Line~\protect\ref{alg:rsm-rel-feat-neigh-labels}-\protect\ref{alg:rsm-rel-feat-neigh-attrs} to the current set of features and renormalize if needed.
\label{alg:rsm-append-rel-features}}

\foreach{{\bf each} $v_i \in \Vu$ in $\mathbf{order}$} \label{alg:rsm-for-each-test-node} \Comment{next test instance} 

\multiline{
Set $\vw_i^{R}$, $\vw_i^{I}$ to $\mathbf{0} = \big[0 \; \cdots \; 0 \big] \in \RR^{\k}$
\label{alg:init-w-and-m-vectors}}

	\parfor[{\bf each} $v_j \in \Vl$] \label{alg:rsm-for-supervised} \Comment{Supervised}
		\State Set $s_{ij}$ to be $\simf\left\langle \vz_i,\vx_j\right\rangle$ \label{alg:rsm-supervised-sim}	
		\State Let $k$ be the index of the class label $y_j$ for $v_j$
		\label{alg:rsm-class-label} 
		
		\If{$v_j \in \N_{h}(v_i)$} \label{alg:rsm-check-neighbor}
			\State \; Update $w^{R}_{ik} \leftarrow  w^{R}_{ik} + p_{ik} \cdot s_{ij}$ \label{alg:rsm-supervised-relational}	
		\Else 
			\; Update $w^{I}_{ik} \leftarrow  w^{I}_{ik} + p_{ik} \cdot s_{ij}$			\label{alg:rsm-supervised-iid} 
		\EndIf
	\parend \label{alg:rsm-for-end-supervised}

	\parfor[{\bf each} $v_j \in V^{u}$] \label{alg:unsupervised-unlabeled} \Comment{Semi-supervised} 
		\State Set $s_{ij}$ to be $\simf\left\langle \vz_i, \vz_j \right\rangle$  \label{alg:rsm-ssl-sim}		
		
		\For{{\bf each} class $k \in \mathcal{C}$} \label{alg:rsm-ssl-for-class} 
			\If{$v_j \in \N_{h}(v_i)$} \label{alg:check-neighbor-unlabeled}
				\State \; $w^{R}_{ik} \leftarrow  w^{R}_{ik} + p_{ik} p_{jk} \cdot s_{ij}$ \label{alg:rsm-unlabeled-neighbor}
			\Else \; $w^{I}_{ik} \leftarrow  w^{I}_{ik} + p_{ik} p_{jk} \cdot s_{ij}$ \label{alg:rsm-unlabeled-independent}
			\EndIf
			
		\EndFor \label{alg:rsm-ssl-for-class-end}
	\parend \label{alg:unsupervised-unlabeled-endfor}
	\State Normalize $\vw^{R}_{i}$ and $\vw^{I}_{i}$ s.t. $\sum{w^{R}_{ik}} = \sum{w^{I}_{ik}} = 1$ \label{alg:normalize-sim-scores}
	
\endforeach \label{alg:rsm-end-for-each-test-node}

\multiline{Update $\vp_i$ via Eq.~\protect\eqref{eq:rsm-update-eq-supervised}, $\;\;\forall\, v_i \in \Vu$
\label{alg:rsm-local-obj}}

\multiline{Compute confidence $\vc$ and assign predictions for top-$k$ most confident
\label{alg:est-confidence}}

\multiline{Include $\mP$, $\mW^{R}$, $\mW^{I}$, and/or $\vc$ as features (if warranted)
\label{alg:meta-features}}

\State Set $\tau \leftarrow \tau + 1$ and renormalize data if needed \label{alg:update-num-iterations-thus-far}

\Until{stopping criterion is reached or $\tau > \Touter$} \label{alg:stopping-criterion}

\smallskip
\ParFor[{\bf each} $v_i \in V^{u}$] \label{alg:final-pred-for}
\multiline{${y}_{i} \; \leftarrow  \arg\max\limits_{ k \in \mathcal{C}} \; P_{ik}$ \Comment{k is the class label}
\label{alg:score-function}}
\EndPar \label{alg:final-pred-endfor}
\end{algorithmic}
\end{spacing}
}
\vspace{0.5mm}
\end{algorithm}
\end{minipage}
\end{center}
\vspace{-4mm}
\end{figure}
}

A general computational framework for $\rsm$ is given in Alg.~\ref{alg:rsm}. The $\rsm$ framework gives rise to a large space of potential relational learning methods due to $\rsm$'s powerful representation and flexibility, \eg, as many of the learning components in Alg.~\ref{alg:rsm} are naturally interchangeable. It is straightforward to learn these from data directly, similar to how hyperparameters of a particular instantiation of $\rsm$ are learned via k-fold cross-validation (CV).

\subsection{Similarity Functions}
\noindent
We define a few parameterized similarity functions.
Note that one may also interpret $\mX\mZ^{\T}$ as a similarity matrix with the dot product as similarity measure.
Thus, $\rsm$ classifies a test point according to a weighted sum of its similarities with the training points.

\subsubsection{Radial Basis Function (RBF).} 
Given vectors $\vx_i$ and $\vz_j$ of length $\f$, the RBF similarity function is:
\[
\simf(\vx_i,\vz_j) = \mathrm{exp}\Big(-\frac{||\vx_i - \vz_j||^{2}_{2}}{2\sigma^2}\Big)
\]
\noindent where the radius of the RBF function is controlled by choice of $\sigma$ (i.e., tightness of the similarity measure).

\subsubsection{Polynomial functions}
Polynomial functions offer another representative similarity measure for vectors of uniform length:
\[
\simf(\mX,\mZ) = ||\mX, \mZ||^q
\]

\subsection{Class Prior Estimation} \label{sec:class-prior-estimation}
\noindent
Let $\mP = \big[P_{ik}\big] =  \big[\, \vp_{1}^{\T} \; \cdots \; \vp_{i}^{\T} \; \cdots \; \big]$
where each $\vp_{i}$ is a $\k$-dimensional row vector.
Given $\n$ training nodes $\Vl$ with known labels, informally, we learn a model which is used as an initial estimate for the nodes with unknown class labels, that is, $\vp_j, \; \forall j=1,...,\m$ and $v_j \in \Vu$.
For simplicity, the experiments in Section~\ref{sec:exp} use the following approach: 
\begin{enumerate}
\itemsep 6pt
\item[$\mathbf{S1}$] Estimate the class-prior probability $\widehat{\Pr}(y)$ from the training data $\{(\vx_i, y_i)\}^{\n}_{i=1}$ as $\widehat{\Pr}(y) = \n_y / \n$ 
and set $\vp_j^{(0)} = \widehat{\Pr}(y) \; : \; \forall \, j=1,\ldots,\m$ 
where $n_y$ is the number of training nodes in $\Vl$ in class $y$.
\item[$\mathbf{S2}$] Estimate $\vp_j, \; \forall v_j \in \Vu$ from training data $\{(\vx_i, y_i) \; | \; \vx_i \in \RR^{\d}, \; y_i \in \{1,2,\ldots,k\}  \}^{\n}_{i=1}$ 
using $\rsm$-iid for simplicity, however, any efficient and accurate approach will suffice.
\item[$\mathbf{S3}$] Iteratively compute new estimates $\vp_j, \; \forall v_j \in \Vu$ using a fast linear-time approach based on belief propagation. This corresponds to a simple collective learning approach that essentially meshes the current probability distribution vector $\vp_j$ for $v_j \in \Vu$ with the estimated probability vectors from the neighbors of $v_j$ denoted $\N(v_j)$ \emph{s.t.} probability vectors for the neighbors $\bar{\vp}_i = \ve^{\T}\mP_{s}/r$ where $\mP_s=\big[ \dotso \; \vp_i^{\T} \; \dotso \big] \in \RR^{r \times \k}$, $\ve^{\T}$ is a vector of all ones, and $r=|\N_h(v_j)|$.
Thus, $\bar{\vp}_i = \ve^{\T}\mP_{s}/r$ is the ``centroid'' of estimates from $v_j$'s neighborhood, and is used to compute $\vp_{j}^{(\tau+1)}$ along with the previous local  estimate of $v_j$.
This is repeated for all vertices across a number of iterations until the estimates converge.
\end{enumerate}
\noindent
This approach differs from existing work that uses a ``local model" to obtain an estimate for each node~\cite{mcdowellCIKM2013,sen2008collective,macskassy2007classification}.
However, a key advantage of $\rsm$ is its flexibility, and as such, other semi-supervised estimation methods are also applicable and may lead to further improvements~\cite{du2014semi}.

\subsection{Update Equations}
\noindent
A key contribution of this work is the proposed relational learning framework that learns from 
\emph{$\iid$} and \emph{relational data} that is both \emph{labeled} and \emph{unlabeled}.
To the best of our knowledge, this work is the first to leverage and investigate using both \emph{labeled} and \emph{unlabeled} relational \emph{and} $\iid$ data in learning.
This leads to four different learning scenarios: 
(a) labeled neighbors, 
(b) unlabeled neighbors, and
(c) labeled and 
(d) unlabeled nodes that are not neighbors.
Thus, $\rsm$ exploits the estimates from labeled and unlabeled $\iid$ and relational data simultaneously (and in a collective fashion) to improve predictive performance.
The update equation used in this paper is simply:
\begin{align} \label{eq:rsm-update-eq-supervised}
\vp_i^{(\tau+1)} = 
\overbrace{
\underbrace{
\vphantom{\Big(}
\; \alpha \vw^{R}_{i} \;}_{\text{Neighbor}} 
+ \;
\underbrace{
\vphantom{\Big(} 
(1-\alpha) \vw^{I}_{i}}_{\text{Non-neighbor}} 
\;
}^{\text{current estimate}}
\; + \;
\overbrace{
\vphantom{\Big(}
\omega
\vp_i^{(\tau)}
}^{\text{previous}}
\end{align}
\noindent
where $\alpha$ is a hyperparameter which satisfies $0 \leq \alpha \leq 1$, and $\vw_{i} \in \RR^{\k}$ is the non-negative vector with $W_{ik} \geq 0$.
Further, $\omega$ determines the influence given to the previous estimate $\vp_i^{(\tau)}$ and $\tau$ is the iteration.

\subsection{Semi-Supervised Learning}
\noindent
Semi-supervised learning (SSL) lies between unsupervised and supervised learning as it uses both unlabeled and labeled data for deciding the class of a node.
Partially labeled data is often found in practice, due to it being costly and time-consuming to obtain labels.
Note $\rsm$ is also flexible for the case where there are very few labeled instances and supports a number of intuitive and principled semi-supervised learning strategies. 
For instance, at each iteration (global update), 
we estimate uncertainty (for each unlabeled instance), 
and predict the labels of the top-$k$ nodes for which we are most certain (min uncertainty).
Notice that Line~\ref{alg:unsupervised-unlabeled}-\ref{alg:unsupervised-unlabeled-endfor} in Alg.~\ref{alg:rsm} leverages unlabeled nodes directly.
Future work will investigate different update equations that leverage these weights differently. For instance, instead of combining these weights into $\vw^{R}_{i}$ and $\vw^{I}_{i}$, we can define different weight vectors to maintain these weights independently of those from Line~\ref{alg:rsm-for-supervised}-\ref{alg:rsm-for-end-supervised}.

\subsection{Classification}
\noindent
We now reinterpret the problem for classification, but it may also be used for regression in a straightforward fashion.
Let $\mX \in \mathbb{R}^{\n \times \f}$ be a matrix where the rows represent training instances and the columns represent features. 
Further, let $\mZ$ be a matrix of test instances, then the class of a single test instance $\vz_j$ is predicted as follows.
First, the similarity of $\vz_j$ with respect to each training example in $\mX$ is computed.
For instance, suppose $\vx_i$ belongs to class $k \in \mathcal{C}$, then $\simf(\vx_i,\vz_j)$ is added to the $k$th element $w_k$ of the weight vector $\vw$.
The similarity of the instances in $\mX$ of class $k$ with respect to the test object $\vz_j$ is formalized as, 
\begin{align}
 w_k = \sum_{\vx_i \in \mathcal{X}_k} \simf(\vx_i,\vz_j)
\end{align}
\noindent where $\mathcal{X}_k$ is the set of training objects from $\mX$ of class $k$. 
Thus $\vw$ is simply,
\begin{align}
\displaystyle
\vw = \left[ \begin{array}{cccc}
\sum\limits_{\vx_a \in \mathcal{X}_1} \simf(\vx_a,\vz_j) &
\ldots &
\sum\limits_{\vx_i \in \mathcal{X}_k} \simf(\vx_i,\vz_j) &
\ldots
\end{array} \right]
\end{align}
After computing $\vw$, then $\vz_j$ is assigned the class that is most similar.
Therefore, we can predict the class of $\vz_j$ using the following decision function:
\begin{align}
\xi(\vz_j) = \argmax\limits_{k \in \mathcal{C}} \;\; w_k
\end{align}
\noindent where $\xi(\cdot)$ is the predicted class.
As an aside, in the case that $\mZ$ is sparse, we compute $\simf(\vx_i,\vz_j)$ fast by hashing its values via a perfect hash function, \ie, and then use this to efficiently test the similarity between only the nonzero elements. 
For real-time systems an even faster approximation may be necessary, in this case, one may compute the centroid from the training examples of each class, and use this as a best estimate.
If there are $\k$ classes, then the complexity for classifying a test point is only $\O(\n\k)$ where $\n$ is the number of columns (features) of $\mX$.

The hyperparameters are learned via k-fold cross-validation on a small fraction of the labeled data.
Other techniques may also be used in a straightforward manner.

\subsection{Confidence} 
\noindent
We also derive a confidence measure based on the estimated weights from $\rsm$.
This confidence measure may be utilized in a variety of ways.
For instance, given the set of nodes for which we have low confidence, we may then hold out on predicting their class labels until the class labels for all the other nodes for which we are confident have been assigned.
We may then use other information to improve our estimates such as biasing the similarity computations using the class labels of the connected nodes (1 or more hops away).

\section{Related Work} \label{sec:related-work}
\noindent
The related work is categorized and reviewed below.

\subsection{Interactive ML}
\noindent
Existing work has focused mainly on traditional machine learning problems that are limited to independent and identically distributed (IID) data~\cite{fails2003interactiveML}.
In addition, Oglic~\etal propose interactive kernel PCA~\cite{oglic2014interactive}.
Kapoor~\etal~\cite{kapoor2012performance} use a human-assisted optimization strategy in the design of multi-class classifiers for $\iid$ data.

\subsection{Relational Learning}
\noindent
Existing methods are inefficient and do not scale to data that is either large or streaming.
In contrast, $\rsm$ is efficient in terms of time and space, while also providing high accuracy for a variety of classification tasks.
Moreover, unlike the previous work, $\rsm$ does not sacrifice speed for accuracy.
Furthermore, most relational learning methods rely on the notion of relational autocorrelation~\cite{introSRL07,rossi2012transforming,wvrn,neville:kdd03}.
An assumption that is often violated in the noisy data found in many real-world settings.
Moreover, in cases where there does not exist high autocorrelation in the data, then relational learning methods may actually be less accurate than traditional $\iid$ machine learning methods.
Additionally, $\irml$ may also be used in combination with relational active learning and active search methods~\cite{wang2013active}.
Relational active learning~\cite{bilgicActiveLearning10}, 
collective classification~\cite{getoor:icdmw07,mcdowell2009cautious},
relational classification~\cite{neville:kdd03,rossi2010drc-modeling,mcgovern2008spatiotemporal}, 
and relational representation learning~\cite{rossi2012transforming} for improving these $\rml$ methods.

\begin{figure}[b!]
\centering
\includegraphics[width=0.98\linewidth]{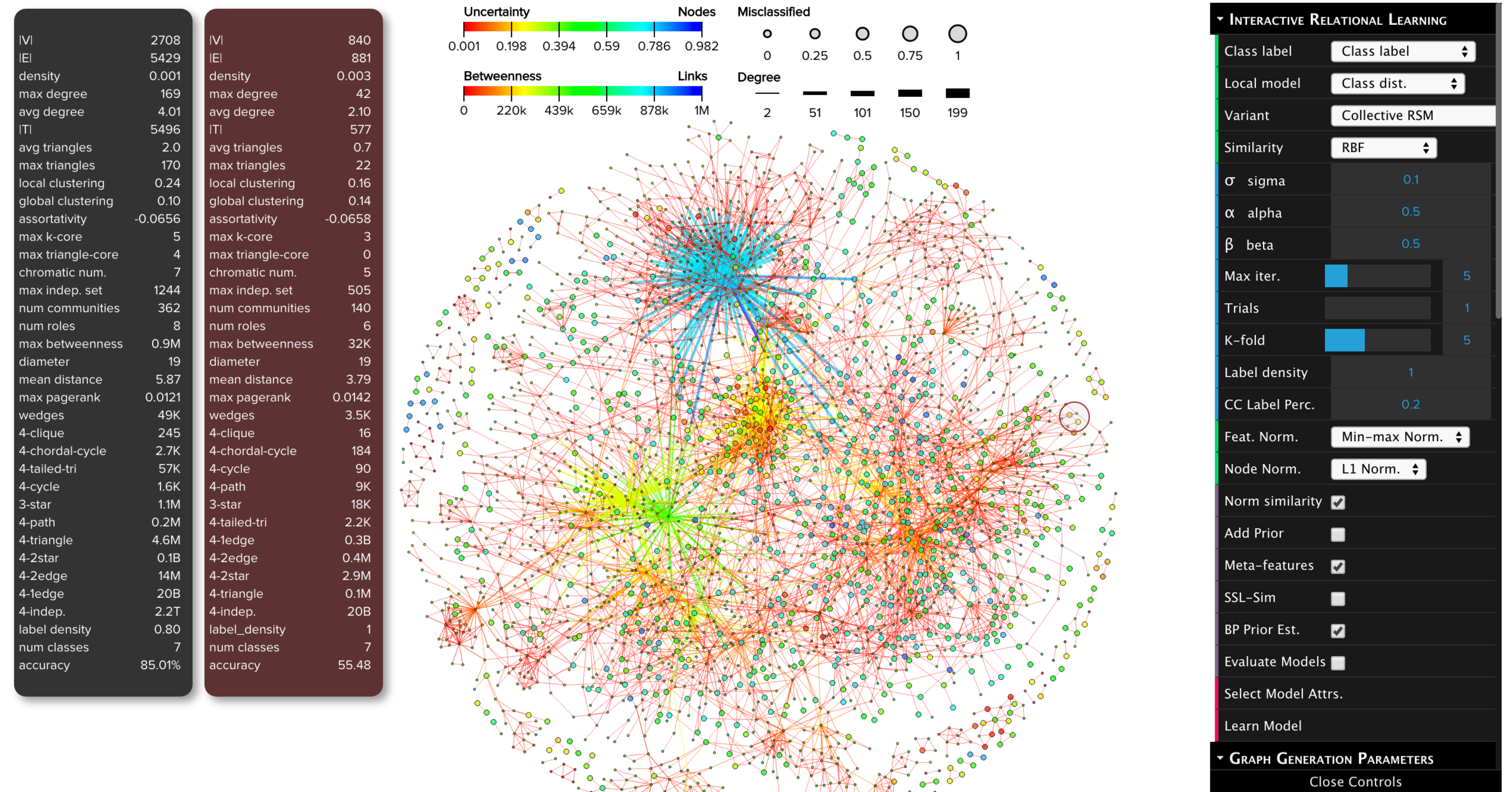}
\vspace{0.1mm}
\caption{Overview of RSM implemented for interactive RML. The visualization above is from $\mathbf{\tt cora}$.}
\label{fig:cora-example}
\end{figure}

\subsection{Visual Analytics}
\noindent
Visual analytics is defined as the science of analytical reasoning supported by interactive visual interfaces~\cite{kielman2009foundations,keim2010mastering,arias2011visual,robertson2009scale}.
Visual analytic methods are becoming increasingly important and have been deployed for numerous real-world applications, including maritime risk assessment~\cite{malik2011visual}, astrophysics~\cite{aragon2009using}, financial planning~\cite{savikhin2011experimental}, law enforcement~\cite{malik2010visual}, and many others~\cite{kim2007visual,andrienko2013visual,sedlmair2011cardiogram}.
Majority of existing work in visual analytics has largely centered around visualization and human computer interaction (HCI) techniques.
However, some work has investigated combining traditional ML methods with visual analytics (e.g., interactive PCA~\cite{jeong2009ipca}, classification via supervised dimension reduction~\cite{choo2010ivisclassifier}).
In contrast, this paper focuses on combining visual analytics with the proposed \emph{{\sf interactive} relational learning} techniques.

\section{Experiments} \label{sec:exp}
\noindent
The experiments are designed to investigate $\rsm$ and its overall effectiveness.
Standard benchmark data for statistical relational learning is used~\cite{mcdowellCIKM2013,wvrn}.
All data has been made available at Network Repository~\cite{nr-sigkdd16}.

\subsubsection{Classification}
\begin{figure}[t!]
\centering
\includegraphics[width=0.98\linewidth]{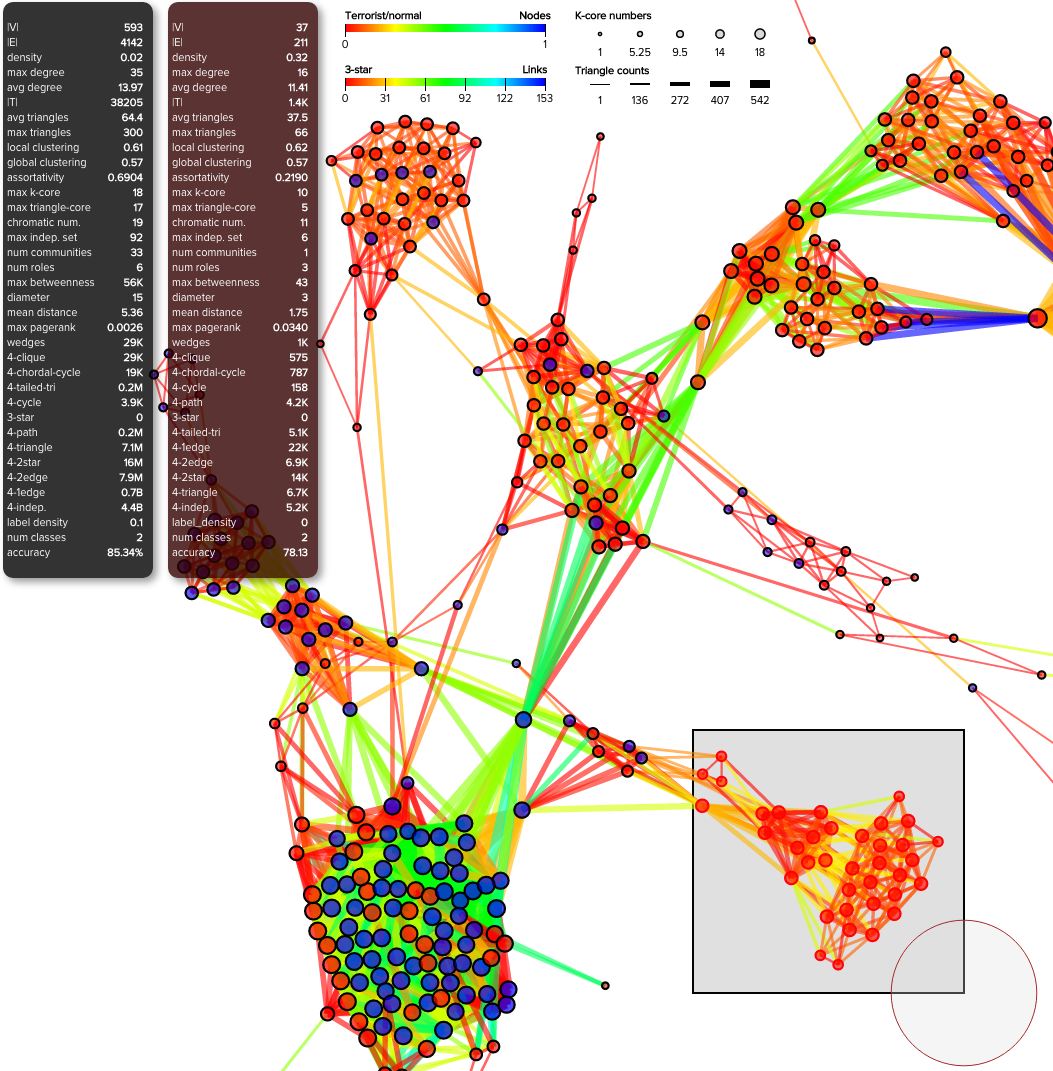}
\vspace{-0.1mm}
\caption{The proposed relational learning framework is not only fast and accurate, but also flexible. 
This gives the user the ability to interactively learn models in real-time and adjust them accordingly, as well as tune parameters, explore/construct features, among many other possibilities.
In the screenshot above (from $\data{soc}{terror}$), the user interactively learns a global model (see the leftmost black side panel for stats and accuracy), and then selects a subgraph $H$ by visually selecting nodes and edges with a simple and intuitive drag-of-the-mouse/gesture.
From this selected subgraph (in real-time), a local RSM model is learned for $H$ and stats and accuracy (using same k-fold CV) are reported in the red side panel on the left side.
Node color represents the class label (terrorist/normal), whereas the link color encodes the number of 3-star graphlet patterns centered at each edge (both can be adapted in real-time by the user).
The weight (or strength) of the nodes and links represent the local max k-core number and triangle counts, respectively.
}
\vspace{-2mm}
\label{fig:soc-terror-overview}
\end{figure}

We investigate the classification performance of the $\rsm$ graph-based learning framework. 
All experiments used $5$-fold cross-validation.
Accuracy is averaged over 20 trials. While $\rsm$ is the first such interactive $\rml$ approach, we nevertheless adapted $\wvrn$~\cite{wvrn,macskassy2007classification} for $\irml$ and used it for comparison.
As an aside, let us note that we also adapted other statistical relational learning (SRL) approaches such as relational dependency networks (RDNs)~\cite{Nevilleunderreview} as well as Probabilistic Soft Logic (PSL)~\cite{psl}, however, these methods were far too slow to support $\irml$.
Furthermore, they were also less accurate than $\rsm$ (and sometimes even $\wvrn$) even despite the significant increase in complexity. 
Initial prior is the class label distribution (overall), and a fast belief-propagation approach is used to obtain an initial estimate.
Results are summarized in Table~\ref{table:data-and-params}. 
Previous work in the related area of relational learning focused mainly on binary classification, however, this work instead focuses on multi-class problems (see Table~\ref{table:data-and-params}).
Note $|C|$ is the number of unique class labels. We also investigated the speed, parallel speedups, and real-time performance capabilities of $\rsm$ for the interactive $\rml$ problem and consistently found it to be extremely fast with real-time response times (in the range of a few ms or less). In addition, Fig.~\ref{fig:soc-terror-overview} demonstrates a few other features supported by $\rsm$.
In particular, $\rsm$ is shown to be fast, parallel, space-efficient, amenable to streaming and dynamic queries/updates, and most importantly, naturally supports real-time interactive learning and inference, and provides rapid immediate (and visual) feedback to the user at real-time interactive response times.
This provides context and allows for quick and intuitive understanding of the intermediate results. 
Upon each change (an additional node/edge is selected visually by the user, ...), a dynamic graph query is issued, and all models, graph properties and statistics are immediately updated and displayed to the user.

\begin{table}[t!]
\caption{Experiments comparing RSM and the adapted iRML-based WVRN. These results demonstrate the effectiveness of RSM. In all cases, RSM outperforms WVRN, and the difference in accuracy is significant. Results that are significant are bolded. Moreover, RSM naturally supports interactive RML, while also significantly more flexible than WVRN ($\eg$, naturally supports graph-based semi-supervised learning).}
\vspace{1mm}
\label{table:data-and-params}
\begin{tabularx}{1.0\linewidth}{r H HHH ccclHH XXHH}
\toprule
\TT \BB & \multicolumn{9}{c}{\bf Hyperparameters} & \multicolumn{4}{c}{{\bf Accuracy}} \TT \BB \\
\TT \BB $\mathbf{Relational}$ $\mathbf{data}$  && $|V|$ & $|E|$ & $|T|$ & $|C|$ & $\sigma$ & $\alpha$ & $\omega$ &&& $\rsm$ & $\wvrn$ & & \\
\midrule

\TTT \BBB $\data{aff}{polbooks}$ 		&&  &  &  & 3 & $0.3$ & $0.7$ & $0.6$ 			&&& $\mathbf{84.73}$ & $74.41$ & & \\
\TTT \BBB $\data{bio}{Gene}$ 			&&  &  & & 2 & $0.2$ & $0.5$ & $0.5$ 			&&& $\mathbf{79.65}$ & $72.41$ & & \\
\TTT \BBB $\data{bio}{Enzymes349}$ 	&& 64 & 118 & $|T|$ & 2 & $0.1$ & $0.25$ & $0$ &&& $\mathbf{77.81}$ & $62.50$ & & \\
\TTT \BBB $\data{aff}{musicGenre}$ 	&& 172 & 212 & 60 & 8 & $0.1$ & $0.75$ & $0.5$ &&& $\mathbf{74.90}$ & $60.59$ & & \\
\bottomrule
\end{tabularx}
\end{table}

\begin{figure}[b!]
\centering
\includegraphics[width=0.65\linewidth]{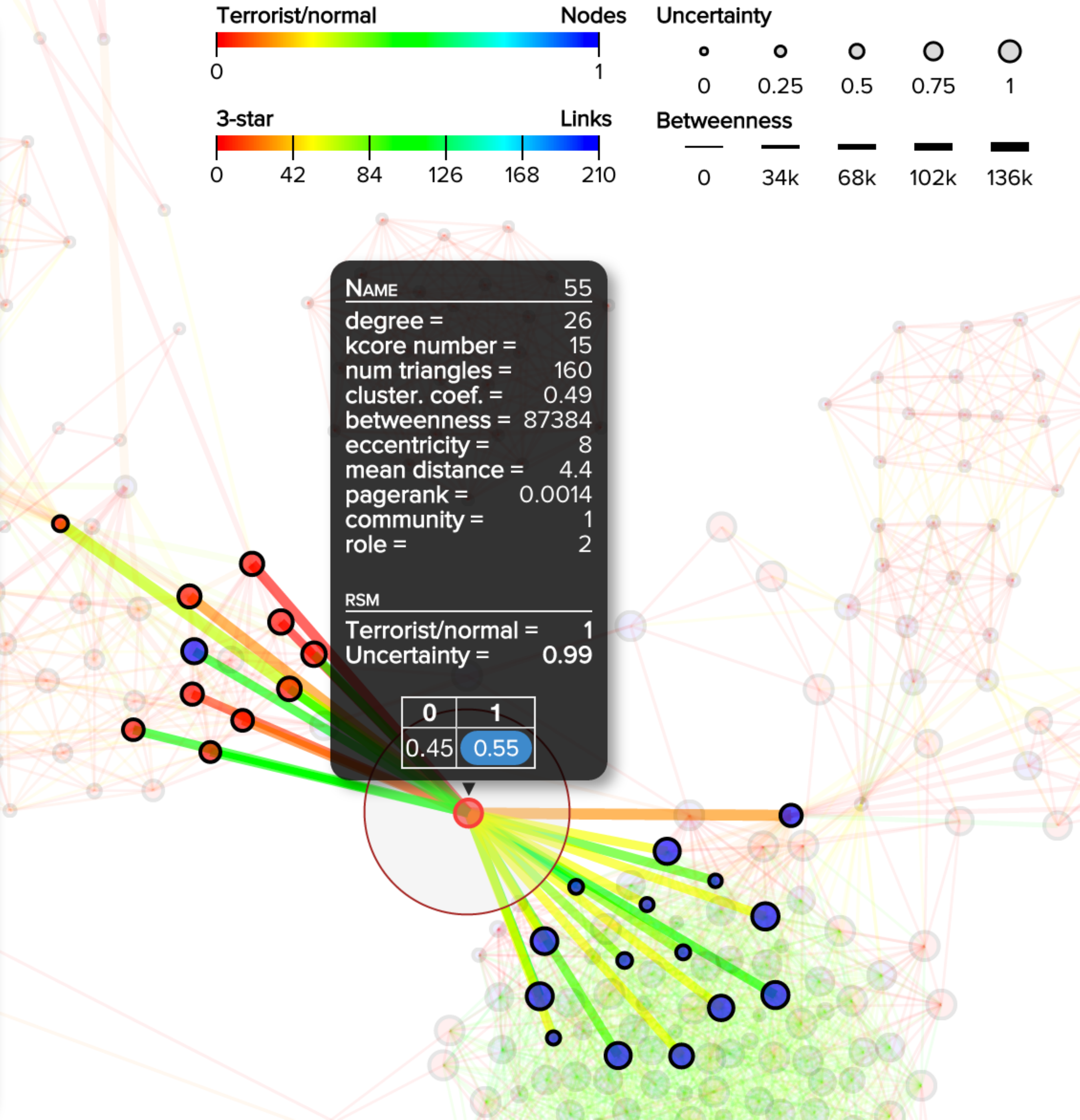}
\caption{Understanding model uncertainty via interactive visual exploration. 
Nodes are colored by class label and weighted (sized) by the uncertainty.
Above RSM enables the user to visually explore the prediction confidence of local nodes, as well as the uncertainty, and other model and graph features via simple and intuitive local dynamic queries (initiated by mousing over a node, or selecting a group of nodes).}
\label{fig:iRML-uncertainty-exploration-soc-terror}
\end{figure}

We also investigated relational learning for sparsely labeled graph data.
For instance, we used $\data{ca}{cora}$ with $10\%$ labels, and only used graph features (that is, no initial attributes were used).
The experiments suggest that other $\rsm$ variants do not work as well due to the high homophily present in $\data{ca}{cora}$.
On the other hand, $\wvrn$ is known to work extremely well in data with precisely these characteristics.
Despite this fact, the graph-based $\rsm$ variant that leverages the $k$-hop neighborhood and the neighbors (graph) features outperforms it significantly.
In particular, it achieves $84.44\%$ accuracy using only the similarity between the $k$-hop neighbors, compared to $78.73\%$ accuracy given by $\wvrn$.
This demonstrates the generality and flexibility of the proposed family of $\rsm$ methods.
While existing relational learning methods have been designed to work for relational data with high homophily,
the experiments demonstrated that $\rsm$ is able to model data with varying levels of homophily and still significantly outperform others that are more specialized and less flexible to handle such data characteristics.

\begin{figure}[h!]
\centering
\includegraphics[width=1.0 \linewidth]{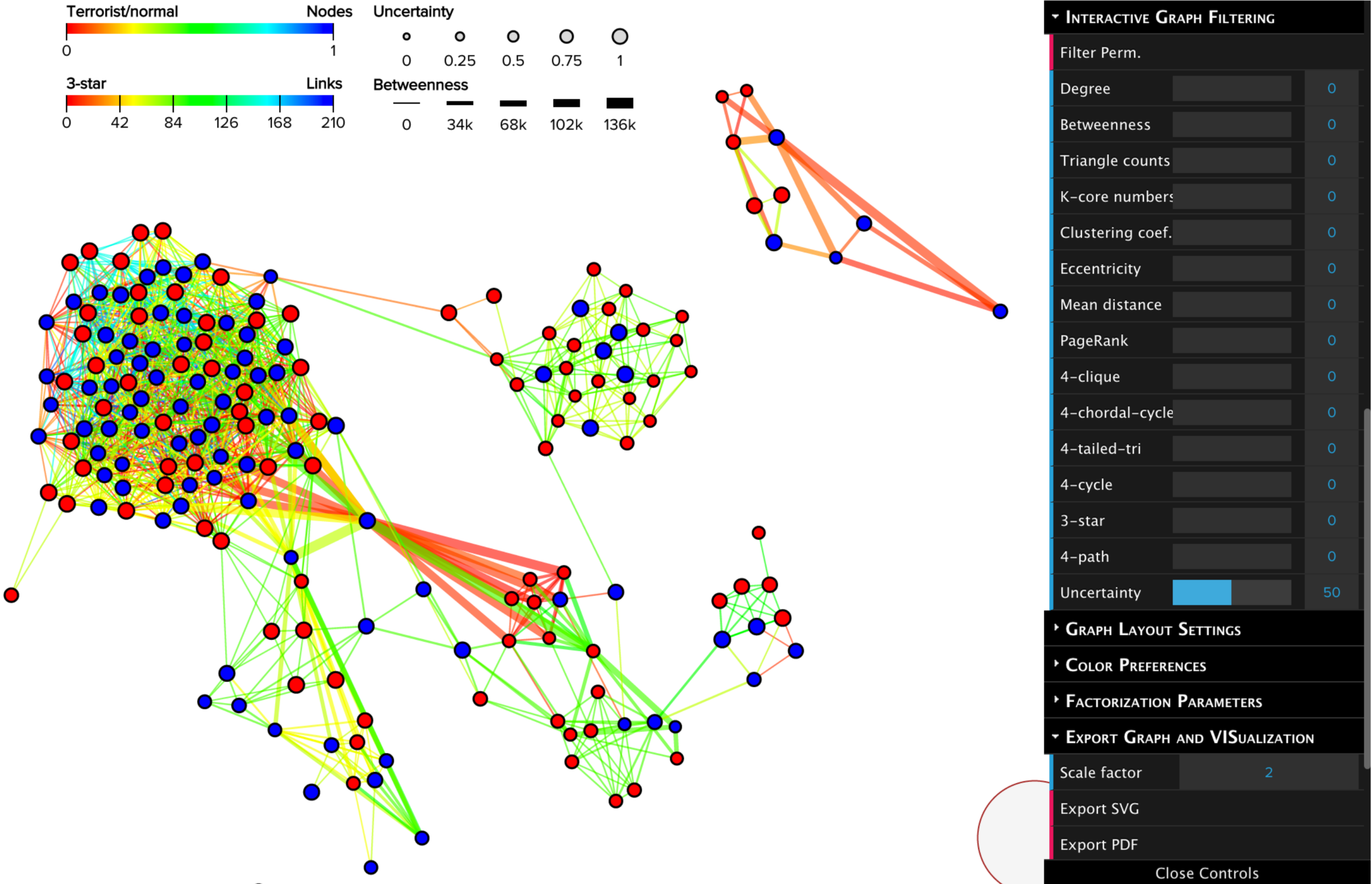}
\caption{Interactive filtering via uncertainty.
All nodes and edges with high confidence are filtered ($\leq 0.50$), leaving only those that are most uncertain.
Relative entropy is used to measure the uncertainty of the learned probability distribution.
Thus uncertainty is bounded between 0 and 1.
}
\label{fig:interactive-filtering-uncertainty-soc-terror}
\end{figure}

As expected, we also find that the weight or contribution assigned to the various types of relational dependencies may significantly impact the predictive quality of the model.
In particular, the classification accuracy of the model is significantly different as the amount of relational information incorporated into the model is varied.
Many results were removed for brevity.

Recent work by McDowell~\etal found that neighbor attributes can improve classification accuracy.
In addition to these types of features, we also investigate using graph features such as the frequency of 4-cliques centered on a vertex (edge) for improving relational learning tasks.
For instance, motifs of size $k$=$4$ and greater are computed using the recent graphlet decomposition algorithm proposed in~\cite{ahmed2015icdm}.
We also investigated other graph features including community label, betweenness, eccentricity, mean distance, k-core numbers, color class, PageRank, triangle counts, clustering coef., degree, roles.
A key finding of these experiments is that the structural features alone may significantly improve predictive performance.
This result implies that the connectivity of the graph gives rise to graph motif patterns that are correlated with the different node labels.
For instance, suppose a web page representing a node in a web graph forms a large number of 3-star motifs, then this node is likely to be a malicious page (spammer).

An overview of the $\irml$ system for $\rsm$ is shown in Figure~\ref{fig:cora-example}.
In that example, we first interactively learn a model, then select the misclassified nodes for further analysis.
The global statistics of the selected subgraph are shown in the right-most panel.
Node color represents the model's uncertainty using an entropy-based measure, 
whereas the size of the node indicates whether it was correctly classified or not.
In Figure~\ref{fig:cora-example}, misclassified nodes are given a larger size so that they can easily be identified for further exploration.
Uncertainty (and the estimated class prob. distribution, statistics, etc.) of a node or set of nodes may also be displayed by selecting or mousing over those nodes of interest (Fig.~\ref{fig:iRML-uncertainty-exploration-soc-terror}).
$\rsm$ also supports interactive real-time visual graph filtering (e.g., remove all uncertain nodes above a user-specified threshold, e.g., see Fig.~\ref{fig:interactive-filtering-uncertainty-soc-terror} and Fig.~\ref{fig:soc-terror-overview}).
In addition, all visualizations are interactive and support brushing, linking, zooming, panning, tooltips, etc (Fig.~\ref{fig:soc-terror-overview}).
Efficient update rules are also derived to avoid relearning the model (after each user interaction/visual query).
For example, after the deletion (or insertion) of a node, we can update the global relational model via a fast localized update.
These local updates enable real-time exploration capabilities by leveraging fast exact or approximate solutions (e.g., to support real-time  interactive queries seamlessly, see Fig.~\ref{fig:soc-terror-overview}).

Many of the components in $\rsm$ may be explored using interactive visualization and analytic techniques, including the attribute to predict, 
initial features to use (non-relational and graph-based features), 
local model for estimation, 
kernel function (RBF, linear, polynomial, etc.),
hyper-parameters (for selected kernel), 
node- and feature-wise normalization scheme (L1, min-max, etc.),   
as well as whether to use semi-supervised learning (SSL), and meta-features (based on current estimates).
For example, see Fig.~\ref{fig:cora-example} (right panel).
Interactive link prediction methods and many other important learning components are also included in our $\irml$ system and can be leveraged directly by $\rsm$ for improving learning and inference.

\subsubsection{Complexity}
\noindent
Given $\n$ training instances, $\m$ test instances, and $\d$ features.
For learning and inference, $\rsm$ takes $\bigO{\n \d}$ time on a single test instance (in the stream), and therefore $\bigO{\n \m \d}$ for all $\m$ test examples.
The complexity for both sparse and dense training feature matrix $\mX \in \mathbb{R}^{\n \times \d}$ is given below.
Consider the case where $\mX$ is sparse and stored as a sparse matrix using compressed sparse column/row format.
Let $\Omega_{\mX}$ denote the number of nonzeros in $\mX$, then the cost of a single test example is $\bigO{|\Omega_{\mX}|}$ linear in the number of nonzeros in $\mX$.
Further, assuming $p$ processing units, then $\bigO{|\Omega_{\mX}|/p}$, and hence is very scalable for real-time systems.
Now suppose $\mX$ is a dense matrix.
Given a dense training set $\mX \in \RR^{\n \times \d}$ (few zeros), it takes $\bigO{\m \d}$ time for each test object, 
and $\bigO{\frac{\m \d}{p}}$ 
assuming $p$ processing units.
Note that the cost may also be significantly reduced by selecting a representative set of training objects using one of the previously proposed methods.

Finally, we discuss the space complexity of the proposed relational learning framework.
Given a single test node to predict, $\rsm$ takes only $\bigO{\k}$ space where $\k$ is the number of classes.
This is of course in addition to the graph and features.
For parallel learning and inference using $p$ workers, the lock-free version of $\rsm$ takes $\bigO{p \k}$ space.
To avoid locks, each worker maintains a $\k$-dimensional vector of similarity scores, which are then combined upon completion.
However, if locks are used then the initial cost of $\bigO{\k}$ holds, but at the expense of runtime.
Now, suppose collective inference is used, then $\rsm$ takes only $\bigO{\n\k}$ space where $\n$ is the number of test nodes with unknown class labels.

\section{Conclusion} \label{sec:conc}
\noindent
This paper proposes a graph-based learning framework called \emph{relational similarity machines} ($\rsm$) for (semi-supervised) learning in large and noisy networks with arbitrary relational autocorrelation. 
Most importantly, $\rsm$ is a fast, accurate, and flexible \emph{relational learning framework} based on the notion of maximizing similarity.
Our approach naturally handles both binary and multi-class classification problems in large attributed and possibly streaming networks.
Despite the importance of relational learning, existing methods are unable to handle relational data with varying levels of relational autocorrelation, and therefore limited in their ability (and utility) in many situations and data characteristics.
In contrast, $\rsm$ naturally handles network data with arbitrary levels of autocorrelation by adjusting a simple hyperparameter.
Furthermore, $\rsm$ has many other advantages over existing methods, including:
(a) it is space- and time-efficient, 
(b) easily parallelizable and thus scalable for large relational data,
(c) naturally allows for varying levels of relational autocorrelation,
(d) flexible as many components are interchangeable (and can be learned from the data directly or fine-tuned by an expert),
(e) amenable to (attributed) graph streams, and 
(f) has excellent accuracy. 
In addition, both node and link classification problems are easily handled in $\rsm$.
Moreover, $\rsm$ is able to leverage node and link attributes simultaneously as well as heterogeneous networks with multiple node and edge types.

{
\fontsize{8}{8.4}\selectfont
\bibliographystyle{abbrv}
\bibliography{paper} 
}
\end{document}